\documentclass{llncs}
\usepackage{amsmath}
\usepackage{amssymb}
\usepackage{graphicx}
\usepackage{bbm}
\usepackage{enumerate}
\usepackage{makecell}
\usepackage[utf8]{inputenc}
\usepackage{color}
\usepackage{fancyhdr} 
\usepackage{float}
\usepackage{caption}
\def\Rand{\textsf{Rand}}
\def\Poss{\textsf{Poss}}

\begin{document}

\title{A Distributional Approach for Soft Clustering Comparison and Evaluation}
\author{Andrea Campagner\inst{1} \and Davide Ciucci \inst{1} \and Thierry Den{\oe}ux\inst{2,3} }
\institute{
University of Milano-Bicocca, viale Sarca 336 -- 20126 Milano, Italy \and
Université de technologie de Compiègne,\\ CNRS, UMR 7253 Heudiasyc, Compiègne, France \and
Institut universitaire de France, Paris, France}
\authorrunning{A. Campagner et al.}

\maketitle

\begin{abstract}
The development of external evaluation criteria for soft clustering (SC) has received limited attention: existing methods do not provide a general approach to extend comparison measures to SC, and are unable to account for the uncertainty represented in the results of SC algorithms. In this article, we propose a general method to address these limitations, grounding on a novel interpretation of SC as distributions over hard clusterings, which we call \emph{distributional measures}.  We provide an in-depth study of complexity- and metric-theoretic properties of the proposed approach, and we describe approximation techniques that can make the calculations tractable. Finally, we illustrate our approach through a simple but illustrative experiment.
\end{abstract}

\section{Introduction}
External clustering evaluation, defined as the act of objectively assessing the quality of a clustering result by means of a comparison between two or more clusterings (one of which is usually assumed to be the \emph{correct one}), is one of the most relevant steps in clustering analysis \cite{xiong2018clustering}.
In the case of hard clustering (HC), where each object is \emph{unambiguously} assigned to a single cluster, several criteria have been considered in the literature \cite{day1981complexity,rand1971objective,vinh2010information}.

By contrast, how to properly evaluate the results of a clustering analysis is much less clear in the case of soft clustering (SC) methods. Several SC methods have been proposed, including rough clustering (RC) \cite{lingras2004interval}, fuzzy clustering (FC) \cite{bezdek1981pattern}, possibilistic clustering (PC) \cite{krishnapuram1993possibilistic} and evidential clustering (EC) \cite{denoeux2004evclus,denoeux2016evidential}.
The development of evaluation measures for SC has largely focused on the extension of common measures \cite{anderson2010comparing}, notably the Rand index, to the setting of FC \cite{campello2007fuzzy,frigui2007clustering,hullermeier2011comparing}, while only recently a formulation of this approach has been introduced for the more general case of EC \cite{denoeux2017evaluating}. Nonetheless, a general approach to extend other comparison measures to SC is still lacking.
Furthermore, existing measures fail to properly distinguish and quantify  different types of uncertainty that can arise in SC \cite{denoeux2017evaluating}, namely: \emph{ambiguity}, i.e., the inability to uniquely assign an object to a single clustering (typical of RC); and \emph{partial assignment}, i.e., the assignment of objects to multiple clusters (typical of FC and PC). 

In this article, we propose a general method to address these limitations, which makes it possible to extend any clustering evaluation to the case of SC. This approach allows us providing a full account of the uncertainty in the two clusterings to be compared. It relies on a novel interpretation of SC as representing distributions over HCs, referred to as \emph{distributional measures}. We provide an in-depth study of the proposed approach, with respect to both computational complexity and metric properties. Furthermore, we describe approximation techniques that can make the approach tractable. Finally, we illustrate the application of the proposed method through a simple but illustrative example involving commonly adopted SC algorithms.

\section{Background and Related Work}
In the following section we provide basic background on clustering (Section \ref{ssec:backclust}) and evaluation measures for SC (Section \ref{ssec:measures}). Background material on metric spaces is presented in Appendix \ref{ssec:backmetric}.

\subsection{Background on Clustering}
\label{ssec:backclust}

Let $X = \{x_1, ..., x_n\}$ be a set of objects.  A  \emph{HC} is an unique assignment of objects in $X$ to \emph{clusters}. Formally, a HC can be represented as a mapping $C : X \mapsto \Omega$, where $\Omega = \{ \omega_1, ..., \omega_k \}$ is a set of clusters. This representation is called \emph{object-based}. An equivalent representation, called \emph{relational representation}, can be obtained by defining a clustering as an equivalence relation $[C] \subseteq X \times X$. Clearly, for the case of HC the two representations are equivalent. Given a clustering $C$, $[C]$ denotes its relational representation. Given two clusterings $C_1, C_2$ we say that they are equivalent iff $[C_1] = [C_2]$; we then write  $C_1 \sim C_2$.


As mentioned in the introduction, in \emph{SC} the unique assignment assumption is relaxed: the intuition is that we allow uncertainty in the cluster assignments. In the most general framework of EC, the uncertainty about cluster assignment is represented  as a Dempster-Shafer mass function. Formally, using the object-based representation, an EC is a set $M = \{ m_x \}_{x \in X}$, where each $m_x$ is a \emph{mass function}: i.e., a function $m_x : 2^\Omega \mapsto [0,1]$ such that $\sum_{A \subseteq \Omega} m_x(A) = 1$.
If the mass functions $m_x$ are \emph{logical}, then the collection $R = \{ m_x \}_{x \in X}$ is said to be a \emph{RC}. A RC can be seen equivalently as a set of HCs \cite{campagner2019orthopartitions}. Namely, a HC $C$ is \emph{compatible} with $R$ if $\forall x \in X$, $C(x) \in R(x)$. Then,  $R$ can be represented by $C(R) = \{ C : C \text{ is compatible with } R\}$.
If all $m_x$ are Bayesian, then the collection $F = \{ m_x \}_{x \in X}$ is a FC. Finally, if all $m_x$ are consonant, then the collection $P = \{ m_x \}_{x \in X}$ is a \emph{PC}. Both FC and PC can be alternatively represented as a collection of cluster membership vectors $F = \{ \mu_x \}_{x \in X}$. In PC it is assumed that $\forall x \in X$, $\max_{\omega \in \Omega} \mu_x(\omega) \leq 1$, while in FC that $\forall x \in X$, $\sum_{\omega \in \Omega} \mu_x(\omega) = 1$.



A relational representation can also be defined  for the case of EC. Let $\Theta = \{s, \neg s\}$ be the frame where $s$ denotes that two objects are in the same cluster, and $\neg s$ the opposite event. Given an EC $M$, the corresponding relational representation can be obtained, for any two objects $x, y \in X$, by combining $m_x, m_y$ using Dempster's rule of combination and then computing the restriction $m^{xy}$ of the resulting mass function to $\Theta$ \cite{denoeux2017evaluating}.

Finally, we note that, if we interpret a SC as describing the uncertainty in regard to an underlying (unknown) HC, then two types of uncertainty can be distinguished. First, \emph{partial assignment}, i.e., the fact that for $\omega_1, \omega_2 \in \Omega$ it may happen that $m_x(\omega_1), m_x(\omega_2) > 0$. Second, \emph{ambiguity}, i.e., the assignment of some mass to non-singleton events, describing our inability to exactly determine to which cluster an object belongs. It is easy to observe that in a FC only partial assignment is relevant, since all the mass is assigned to the singletons, while in the case of RC, only ambiguity is present. By contrast,  the EC formalism is flexible enough to represent both types of uncertainty.

\subsection{Clustering Comparison Measures}
\label{ssec:measures}

Several measures have been defined to compare clusterings. 
Given two HCs $C_1, C_2$, a commonly adopted approach to compare them is to evaluate the number of object pairs   $x, y \in X$ on which they agree. Formally, the Rand index can be defined as:
\begin{align}
    \Rand(C_1, C_2) &= \frac{|\{(x,y) \in X^2 : (x,y) \in ([C_1] \cap [C_2]) \cup ([C_1]^c \cap [C_2]^c)\}|}{|X|^2}.
\end{align}
It is easy to show that the Rand index is a similarity on HCs.

Several extensions of the Rand index to the setting of SC have been considered. In the case of FC and PC, Campello et al. \cite{campello2007fuzzy} and Frigui et al. \cite{frigui2007clustering} proposed to use a t-norm $\wedge$ and a t-conorm $\vee$ in place of classical set operators. This setting was later generalized to the setting of RC \cite{depaolini2018external}, by means of a transformation from RC into FC. 
A further generalization of the Rand index was proposed by H\"ullermeier et al. \cite{hullermeier2011comparing}, in the setting of FC, and Den{\oe}ux et al. \cite{denoeux2017evaluating}, in the more general setting of EC. This approach is based on the use of a normalized metric $d_M$  on mass functions, that is then extended to compare ECs as 
\begin{equation}
\label{eq:randevid}
   \Rand_{E}(M_1, M_2) = \frac{2}{n(n-1)}\sum_{x \neq y \in X} 1 - d_M(m_1^{xy}, m_2^{xy}).
\end{equation}
 The authors of \cite{denoeux2017evaluating,hullermeier2011comparing} note that, while the measure they propose is a pseudo-similarity, it is not completely satisfactory for the comparison of ECs as it does not distinguish between \emph{partial assignment} and  \emph{ambiguity}.

Another commonly adopted approach for the definition of clustering evaluation measure grounds on information theory. Let $C_1, C_2$ two HCs.
The mutual information is defined as
\begin{equation}
    MI(C_1, C_2) = \sum_{\omega_i^1 \in \Omega_1} \sum_{\omega_j^2 \in \Omega_2} p_{ij} \log \frac{p_{ij}}{p_i^1 \cdot p_j^2},
\end{equation}
 where $p_i^1 = |\{ x \in X : C_1(x) = \omega_i^1 \}|/|X|$, and similarly for each $\omega_i^2 \in \Omega_2$, while $p_{ij} = |\{ x \in X: C_1(x) = \omega_i^1 \text{ and } C_2(x) = \omega_j^2 \}|/|X|$. The mutual information is a similarity over HCs. This measure has been extended to the case of RC in \cite{campagner2019orthopartitions}, by representing RCs as collections of compatible HCs. To our knowledge, no proposal to extend these metrics to the more general case of EC can be found in the literature.

Finally, another comparison approach grounds on the notion of \emph{edit distance} between partitions. The \emph{partition distance} \cite{day1981complexity} which, for two HCs $C_1, C_2$, is defined as the minimum number of objects to be moved to transform $C_1$ into $C_2$ (or, equivalently, $C_2$ into $C_1$), can be computed as
\begin{equation}
    d_{\pi}(C_1, C_2) = \frac{1}{|X| - 1} \min_w \frac{1}{2}\sum_i |\omega_i^1 \Delta \omega_{w(i)}^2 |,
\end{equation}
where $w$ is a permutation function, and $\Delta$ is the symmetric difference operator. It is easy to note that $d_\pi$ is a normalized metric.
An extension of the partition distance to the case of FC was proposed by Zhou \cite{zhou2005new}. The obtained measure is a proper generalization of the partition distance and is a metric. 
As for the mutual information, to our knowledge, the extension of the partition distance to rough and EC has not been considered in the literature.

\section{A General Framework for Soft Clustering Evaluation Measures}
\label{sec:method}
As shown in the previous section, most of the research on clustering comparison measures for SC has focused on the analysis of some specific indices, while a general methodology for obtaining such measures is still missing. Furthermore, as noted in \cite{hullermeier2011comparing,denoeux2017evaluating}, most of the existing methods fail to satisfy reasonable metric properties and can thus hardly be used for the objective comparison of SCs. Notably, also the more principled approaches \cite{denoeux2017evaluating} can have some drawbacks, such as the inability to properly distinguish between different types of uncertainty arising in SC.
In this section, we propose an approach that attempts to address these limitations, based on the representation of a SC as a distribution over HCs. 


\subsection{Distribution-based Representation of Soft Clustering}
\label{ssec:distribrepr}

As shown in Section \ref{ssec:backclust},  a RC $R$ can be represented as a set $C(R)$ of HCs $C$. Based on this observation, we extend this representation to general SCs. 
Formally, given an EC $M$, we consider the following probability distribution over RCs:
\begin{equation}
\label{eq:distribevid}
    m_M(R) = \prod_{x\in X} m_x(R(x)),
\end{equation}
which can also be seen as a Dempster-Shafer mass function over HCs. Given an EC $M$ and its distribution-based representation $m_M$, we denote with $\mathcal{F}(M)$ the collection of focal RCs of $m_M$, that is $\mathcal{F}(M) = \{ R : m_M(R) > 0 \}$.

The distribution-based representation for RC, FC and PC can then be obtained as a special case of Eq \eqref{eq:distribevid}. Indeed, in the case of RC, $m_M$ is logical (i.e. $|\mathcal{F}(m_M)| = 1$). In the case of a FC  $F = \{ \mu_x \}_x$, where $\mu_x : \mathcal{C} \mapsto [0,1]$ is a probability distribution, the focal RCs are all singletons (i.e., HCs), thus we can define $Pr_F(C) = \prod_{x \in X} \mu_x(C(x))$
Finally, given a PC $P$ and a t-norm $\wedge$, we can view $P$ as a possibility distribution over HCs $\Poss_P(C) = \bigwedge_{x \in X} \mu_x(C(x))$.
The possibility distribution $\Poss_P$ can equivalently be represented as a consonant mass function, i.e., a mass function for which the focal RCs are nested. Note that, when $\wedge$ is the product t-norm, we recover the case of FC.

 \subsection{Distributional Measures}
 \label{ssec:distribdistances}
 Let $d$ be a normalized metric on HCs. Since, as shown in the previous section, any SC can be seen as a distribution over HCs, an intuitive approach would be to extend $d$ to a distribution-valued function, providing a quantification of the belief about the real value of the evaluation measure. The intuition behind this idea is based on the definition of SC as representing a clustering with some uncertainty affecting our knowledge with respect to the assignment of objects to clusters. Thus, it is natural to require that an evaluation measure for SC should transfer this uncertainty to the possible outcomes of the evaluation.

Therefore, intuitively, a measure over RCs would provide, given two RCs $R_1, R_2$, a set of values, representing all possible distances between HCs compatible with $R_1, R_2$. More generally, a measure over ECs would provide a mass function over possible values of $d$. Formally, we define the \emph{distributional measure}, based on $d$, between two RCs (resp., PCs, ECs) as, respectively:
 \begin{align}
     d_R(R_1, R_2) &= \{ d(C_1, C_2) : C_1 \in C(R_1) \text{ and } C_2 \in C(R_2) \}\\
     \forall v \in \mathbbm{R}, \; d_P(P_1, P_2)(v) &= \bigvee_{C_1, C_2: d(C_1, C_2) = v} \Poss_{P_1}(C_1) \wedge \Poss_{P_2}(C_2)\\
     \forall V \in 2^\mathbbm{R}, \; d_E(M_1, M_2)(V) &= \sum_{R_1, R_2 : d_R(R_1, R_2) = V} m_{M_1}(R_1) \cdot m_{M_2}(R_2)
 \end{align}
 where $\wedge, \vee$ are a t-norm and the corresponding dual s-conorm.
 It is easy to observe that $d_E$ is a generalization of $d_R$ and $d_P$. For ease of notation, we distinguish the case of $d_P$ where $\wedge = \otimes_P, \vee = \oplus_P$ (resp., the product t-norm and the corresponding t-conorm) and we denote it as $d_F$, since it can be applied directly to the case of FC.
 
 
 
 Intuitively, $d_E(M_1, M_2) = V$ can be interpreted as the evidence supporting the statement ``The distance between the two real HCs underlying $M_1, M_2$ is within $V$''. Therefore, $d_E$ provides a complete representation of the possible distance values that arise when comparing HCs compatible with $M_1, M_2$.
  Though, clearly, $d_E$ is not a metric, we can see that it satisfies the  properties stated in the following theorem.
  
 \begin{theorem}
Function $d_E$ satisfies (M3) (see Appendix \ref{ssec:backmetric}). $d_E(M_1, M_2)(0) = 1$ iff there exists a HC $C$ s.t. for $i=1, 2$, $R \in \mathcal{F}(M_i) \implies \forall C' \in C(R), C' \sim C$.
 \end{theorem}
As a consequence of the previous result, the value $d_E(M_1, M_2)(0)$ can be interpreted as the evidence that the unknown HCs corresponding to $M_1$ and $M_2$ are the same (have a distance equal to 0). Indeed, $d_E$ assigns full evidence to value 0, if and only if $M_1, M_2$ are totally compatible. In particular, simple equality between $M_1, M_2$ does not suffice to obtain $d_E(M_1, M_2)(0) = 1$, unless $M_1, M_2$ are HCs.
 
In regard to computational complexity, it is easy to show that computing $d_R$ (resp., $d_F$, $d_P$, $d_E$) is computationally easy w.r.t. the size of the distribution-based representation introduced in the previous section, while it is intractable w.r.t. the size of the object-based representation:
 \begin{theorem}
 The problem of computing $d_R$ (resp., $d_F$, $d_P$, $d_E$)  has complexity $O(k^m)$, where $m = |\{ x \in X: |R(x)| \neq 1\}|$ and $k = |\Omega|$ is the number of clusters. More precisely, $d_R$ can be computed in constant amortized time, while $d_F, d_P, d_E$ can be computed in at most linear amortized time.
 \end{theorem}
 
 \subsubsection{Interval representation}
 \label{ssec:interval}
A possible solution to the intractability of computing the distributional measures would be to consider a compact representation of the latter. For the case of RC, $d_R$ could be summarized as the interval:
$$\langle d_R^l, d_R^u \rangle (R_1, R_2) = \langle \min \{v  \in \mathbbm{R}: v \in d_R(R_1, R_2) \}, \max \{v  \in \mathbbm{R}: v \in d_R(R_1, R_2) \} \rangle.$$
We note that this definition satisfies the following properties:
\begin{proposition}
\label{prop:distribrough}
Let $R_1, R_2$ be two RCs. Then, $1 - d_R^l$ is a consistency: in particular $d_R^l = 0$ iff $C(R_1) \cap C(R_2) \neq \emptyset$. By contrast, $d_R^u$ is a meta-metric: in particular, $d_R^u = 0$ iff $R_1 = R_2$ and $|C(R_1)| = 1$, that is iff $R_1 = R_2$ is a HC.
\end{proposition}
\begin{corollary}
\label{corol:roughdistribmetric}
$d_R^u$ is a metric iff either $R_1, R_2$ is a HC.
\end{corollary}

As a result of the previous corollary, in the special case where the aim is to evaluate a RC $R$ with respect to a HC $C$ representing the ground truth, then $d_R^u$ is guaranteed to be a metric. Nonetheless, it is easy to observe that computing $\langle d_R^l, d_R^u \rangle$ is still computationally hard:
\begin{theorem}
\label{prop:hardnessrough}
Let $R_1, R_2$ be two RCs, represented through the object-based representation.
    Then, the problem of computing $\langle d_R^l, d_R^u \rangle$ is \textsc{NP-HARD}\footnote{We note that the problem is trivially in \textsc{P} if instead $R_1, R_2$ are represented through the distribution-based representation.}.
\end{theorem}

 
For the case of FC, PC and, more generally, EC, a possible approach to obtain a similar summarization would be to apply a decision rule to transform the distribution-valued $d_F, d_P, d_E$ into simpler indices \cite{denoeux2019decision}. An example of this approach would be to compute the following lower and upper expectations:

\begin{align}
    \underline{E}(d_E)(M_1, M_2) = \sum_{V \subseteq 2^\mathbbm{R}} d_E(M_1, M_2)(V) \min_{d \in V} d = E(d_R^l),\\
    \overline{E}(d_E)(M_1, M_2) = \sum_{V \subseteq 2^\mathbbm{R}} d_E(M_1, M_2)(V) \max_{d \in V} d = E(d_R^u).
\end{align}
If $M_1, M_2$ are two FCs we obtain that $\underline{E}(d_E) = \overline{E}(d_E) = E(d_F)$. Similarly to the case of RC, it is easy to show that the following properties hold:

\begin{theorem}
    Let $M_1, M_2$ be two ECs. Then $\overline{E}(d_E)$ is a meta-metric. Furthermore, $\underline{E}(d_E)$ satisfies only (M1b) and (M2) (see Appendix \ref{ssec:backmetric}). In particular:
    \begin{itemize}
        \item If $M_1, M_2$ is a HC, then $1 - \underline{E}(d_E)$ is a consistency and $\overline{E}(d_E)$ is a metric;
        \item If $F_1, F_2$ are two FCs, then $E(d_F)$ is a meta-metric.
    \end{itemize}
\end{theorem}

From the computational complexity point of view, computing $\underline{E}(d_E), \overline{E}(d_E)$ is at least as hard as computing $\langle d_R^l, d_R^u \rangle$. However, for the case of FC, it is easy to show that for certain base distances $d$, $E(d_F)$ can be computed efficiently:
\begin{proposition}
\label{prop:effrandfuzzy}
    Let $d = 1 - \Rand$. Then, $E(d_F)$ can be computed in time $O(n^2)$.
\end{proposition}
We leave the problem of characterizing the general complexity of computing $\underline{E}(d_E), \overline{E}(d_E)$ as open problem.

 

\subsection{Approximation Methods}
\label{ssec:approximation}
In the previous section we proposed  distributional measures as a general approach to extend any HC comparison measure to a SC comparison measure. Nonetheless, the computation of these distributional measures is, in general, intractable. For this reason, in this section, we introduce some approximation methods and algorithms, based on a sampling approach, which can be applied to any base distance between HCs. 

 We start with the case of the summarized representation of $d_R$, that is with $d_R^l, d_R^u$. Given two RCs $R_1, R_2$, we draw $s$ samples $(C_1^1, C_2^1), \ldots, (C_1^s, C_2^s)$ uniformly from $C(R_1), C(R_2)$. Then, we can approximate $d_R^l$ and $d_R^u$ as, respectively,  $\hat{d}_R^l = \min_{i \in \{1,\ldots,s\}} d(C_1^i, C_2^i)$ and $\hat{d}_R^u = \max_{i \in \{1,\ldots,s\}} d(C_1^i, C_2^i)$. Clearly, the following result holds:

\begin{proposition}
The following bounds hold for any $\epsilon > 0$:
\begin{align}
\label{eq:approxroughdistrib}
   Pr(d_R^u - \hat{d}_R^u > \epsilon) &\leq F(d_R^u - \epsilon)^s \\
   Pr(\hat{d}_R^l - d_R^l > \epsilon) &\leq 1 - \left(1 - F(\epsilon - d_R^l)\right)^s
\end{align}
where $F$ is the cumulative distribution function (CDF) of the probability distribution $p_R$ defined as $p_R(t) = \frac{|\{ C_1 \in C(R_1), C_2 \in C(R_2) : d(C_1, C_2) = t \}|}{|d_R(R_1, R_2)|}$.
\end{proposition}

Since for each $\epsilon$, the quantity $F(d_R^u - \epsilon)$ (resp., $F(\epsilon - d_R^u)$) is stricly less than 1, it holds that $Pr(d_R^u - \hat{d}_R^u > \epsilon)$ (resp., $Pr(\hat{d}_R^l - d_R^l > \epsilon)$) decays exponentially w.r.t. the size of the sample $s$.
However, we note that the quality of the previously described approximation method largely depends on $d_R$. In particular,  the convergence in Eq \eqref{eq:approxroughdistrib} is influenced by the \emph{tailedness} of $p_R$: the heavier the tails of $p_R$, the lower the approximation error. 



For the case of FC, if we use the expected value $E(d_F)$ to summarize $d_F$ and we use a sampling procedure to estimate $E(d_F)$ as $\hat{d}_F$ then we can obtain a tail bound by applying Hoeffding's inequality:
\begin{proposition}
Assume that $d$ is a normalized metric on HCs. Then:
\begin{equation}
\label{eq:approxfuzzydistrib}
    Pr(|\hat{d}_F - E(d_F)| \geq \epsilon) \leq 2 e^{-2s\epsilon^2}
\end{equation}
Hence, the deviation between the empirical mean $\hat{d}_F$ and $E(d_F)$ has exponential decay in the size of the sample $s$.
\end{proposition}

Combining Eqs \eqref{eq:approxroughdistrib} and \eqref{eq:approxfuzzydistrib}, a similar result holds also for $d_E$:
\begin{proposition}
Assume that $d$ is a normalized metric on HCs. Let $\hat{d}_E^l, \hat{d}_E^u$ be the sample estimates of $\underline{E}(d_E), \overline{E}(d_E)$. Then:
\begin{align}
    Pr(|\hat{d}_E^l - \underline{E}(d_E)| \geq \epsilon) &\leq 2 e^{-2s\epsilon^2},\quad
    Pr(|\hat{d}_E^u - \overline{E}(d_E)| \geq \epsilon) &\leq 2 e^{-2s\epsilon^2}
\end{align}
\end{proposition}
Given two ECs $M_1, M_2$, the previous estimate requires that $\hat{d}_E^l, \hat{d}_E^u$ are computed by sampling pairs $R_1, R_2$ of RCs from the distributions $m_{M_1}, m_{M_2}$ and then computing the exact values of $d_R^l(R_1, R_2), d_R^u(R_1, R_2)$. As a consequence of Proposition \ref{prop:hardnessrough}, this may not be feasible when $|X|$ is large. In such cases, a possible solution would be to compute $\hat{d}_E^l, \hat{d}_E^u$ by means of nested sampling (i.e, first we sample a RC $R$ from $m_M$, then we sample a HC $C$ from $C(R)$). In this case, however, one should expect a larger approximation error. 

Finally, we note that all the above mentioned sampling-based approximation methods can easily be implemented in time complexity $O(n^2s)$. 

\
 
 \section{Illustrative Experiment}
 
In this section, we illustrate the use of the proposed metrics using the Anderson's Iris dataset. This latter is a small-scale benchmark problem comprising 150 objects, four numerical features and three perfectly balanced classes. We selected this dataset as, the three classes being approximately linearly separable, it can be expected that any SC algorithm would give as output a clustering close to being an HC. As a consequence of Theorem \ref{prop:hardnessrough}, this is a necessary condition for efficient computation of the exact versions of the distributional measures.

 We considered five different clustering algorithms, namely: k-means (KM), 
 rough k-means (RKM) \cite{peters2014rough}, fuzzy c-means (FCM) \cite{bezdek1981pattern}, possibilistic c-means (PCM) \cite{krishnapuram1993possibilistic} and evidential c-means (ECM) \cite{masson08}.
In order to reduce the complexity of computing the distributional measures, we set the algorithm hyper-parameters  to obtain clusterings close to being hard. In particular, for RKM we set $\epsilon=1.1$, for FCM and PCM we set $m = 5$, and for ECM we set $\delta=10, \beta=5, \alpha=5$.
  The output of each algorithm was compared with the ground truth labeling of the iris dataset. We considered, in particular, the distributional generalizations of the Rand index (D-RI) and the partition distance (D-PD), as well as their sampling-based approximations (S-RI, S-PD).
 All code was implemented in Python (v. 3.8.8), using scikit-learn (v. 0.24.1), numpy (v. 1.20.1) and scipy (v. 1.6.2). 
 
 \begin{center}
     \footnotesize
     \captionof{table}{Results of the experiment. For the Rand index higher is better, while for the partition distance lower is better.\label{tab:results}}
     
     \begin{tabular}{c|c|c|c|c|c}
          Metric & KM & RKM & FCM & PCM & ECM  \\\hline
          D-RI & \makecell{0.877\\(0.034s)} & \makecell{(0.874, 0.886)\\(0.802s)} & \makecell{0.876\\(784.388s)} & \makecell{(0.839, 0.941)\\(979.053s)} & \makecell{(0.781, 0.944)\\(1394.69s)} \\\hline
          S-RI & - & \makecell{(0.874, 0.886)\\(0.429s)} & \makecell{0.876\\(11.266s)} & \makecell{(0.860, 0.927)\\(19.848s)} & \makecell{(0.681, 0.819)\\(19.845s)}\\\hline
          
          D-PD & \makecell{0.111\\(0.031s)} & \makecell{(0.099, 0.113)\\(0.803s)} & \makecell{0.112\\(184.707s)} & \makecell{(0.033, 0.122)\\(224.31s)} & \makecell{(0.041, 0.229)\\(431.57s)} \\\hline
          S-PD & - & \makecell{(0.100, 0.113)\\(0.202s)} & \makecell{0.112\\(11.391s)} & \makecell{(0.072, 0.103)\\(13.739)} & \makecell{(0.154, 0.209)\\(16.424s)} \\\hline
     \end{tabular}
 \end{center}
 
 The results of the experiment are reported in Table \ref{tab:results}, in terms of the metrics values as well as  running time (in seconds). 
  As far as running time is concerned, we can observe that the cost of computing the exact versions of the proposed measures sharply increases when considering more general SC algorithms. Indeed, the running times of D-RI and D-PD for ECM were approximately twice those of FCM and PCM. On the other hand, the differences in running time for the approximation algorithms were much smaller, and indeed the running times for FCM, PCM and ECM were similar.
 
 In terms of approximation quality, even though for RKM and FCM there were no differences between the approximated and exact results, this was not the case for PCM and ECM. In particular, we note that the sampling-based approximation algorithm systematically underestimated the uncertainty in clustering comparison results, by producing intervals that were  narrower than the exact ones. Nonetheless, we note that the approximation methods provided results that were aligned with the exact ones, with smaller values according to one method associated to smaller values according to the other one.

\section{Conclusion}
In this article we proposed a general framework for extending clustering comparison measures from HC to EC (hence, as special cases, also to RC, FC and PC), that we called \emph{distributional measures}. We studied the metric- and complexity-theoretic properties of the proposed measures and, since a major limitation of the proposed approaches lies in their high computational complexity, we also proposed some strategies for approximation based on sampling . 
Finally, we illustrated the application of the proposed methods through a simple experiment.

We believe that this article could provide a first step toward the development of general and principled approaches for the comparison of SC algorithms. For this reason, we deem the following problems to be worthy of further investigation: 1) Generalizing Proposition \ref{prop:effrandfuzzy} to other base distance measures, and determining whether this result can be extended to EC; 2) Designing more refined sampling approaches that can be used to correct the   uncertainty underestimation that we observed in the experiments.

\clearpage
\appendix

\section{Appendix: Background on Metric Spaces}
\label{ssec:backmetric}
 Let $X$ be a set. A \emph{metric} over $X$ is a function $d : X \times X \mapsto \mathbbm{R}$ such that the following condition holds: (M1) $d(x, x) = 0$; (M2) $\forall x \neq y \in X$ $d(x,y) > 0$; (M3) $\forall x, y \in X$ $d(x, y) = d(y, x)$; (M4) $\forall x, y, z \in X$ $d(x, z) \leq d(x,y) + d(y,z)$.
We say that $d$ is normalized if $max_{x, y \in X} d(x,y) = 1$. If $d$ is a normalized metric, then its dual $s = 1 - d$ is called a \emph{similarity} over $X$.

We say that $d: X \times X \mapsto \mathbbm{R}$ is: a \emph{pseudo-metric} iff it satisfies (M1), (M3) and (M4); a \emph{semi-metric} iff if satisfies (M1), (M2) and (M3);  a \emph{meta-metric} iff it satisfies (M2), (M3), (M4) and $\text{(M1b)} \enspace \forall x, y \in X$ if $d(x, y) = 0$, then $x = y$. A \emph{consistency} is a normalized semi-pseudo similarity.

Given a metric $d$ over $X$, the \emph{Hausdorff metric} $d_H : 2^X \times 2^X \mapsto \mathbbm{R}$, based on $d$, is defined as $d_H(A,B) = \max \{ \max_{a \in A} d(a,B) , \max_{b \in B} d(A, b) \}$,
where $d(a,B) = \min_{b \in B} d(a,b)$ and $d(A, b) = min_{a \in A} d(a,b)$. 

\section{Appendix: Proofs}
\begin{proof}[Theorem 2]
For the first part of the result, we note that, given a RC $R$, the size of $C(R)$ is in the worst case exponential in the size of $R$ \cite{campagner2019orthopartitions}. Thus, the statement for $d_R$ follows. Similar considerations can be applied to the cases of $d_F$, $d_P$ and $d_E$. We note that in all cases the size of the distribution-based representation is on the order of $O(k^m)$, therefore for this latter representation the problem of computing $d_R$ (resp., $d_F$, $d_P$, $d_E$) has at most quadratic complexity.
 
For the second part, we note that partitions of a set can be enumerated in constant amortized time \cite{stamatelatos2021lexicographic}, therefore the same holds for computing $d_R$. For the case of $d_F, d_P, d_E$, the same algorithm used for computing $d_R$ can be used: however, for each RC we need to determine its mass. This can be easily performed in time $O(nk)$, which, assuming $n > k$, is in linear-time.
\end{proof}

\begin{proof}[Proposition 1]
Clearly, $d_R^l$ and $d_R^u$ satisfy (M3). Furthermore, it is easy to show that $d_R^l$ satisfies also (M1), while it fails to satisfy (M2). For the case of (M4), consider three RCs $R_1, R_2, R_3$ s.t. $C(R_1) \cap C(R_2) \neq \emptyset$, $C(R_2) \cap C(R_3) \neq \emptyset$, while $C(R_1) \cap C(R_3) = \emptyset$. Thus, $d_R^l$ does not satisfy (M4). Claims for $d_R^u$ similarly follow.
\end{proof}

\begin{proof}[Corollary 1]
One side of the implication directly derives from the observation that in case either $R_1, R_2$ is a HC, $d_R^u$ coincides with the Hausdorff distance between $C(R_1)$ and $C(R_2)$. On the other hand, assume $d_R^u$ is a metric. Then, in particular $d_R^u(R,R) = 0$. This implies that $d_R(R,R) = \{ 0\} \implies |C(R)| = 1$.
\end{proof}

\begin{proof}[Theorem 4]
Clearly, $\overline{E}(d_E)$ satisfies (M2), (M3), (M4). Furthermore, $\overline{E}(d_E) = 0$ iff $M_1 = M_2$ and $M_1, M_2$ are HCs.
For $\underline{E}(d_E)$, this derives from Proposition \ref{prop:distribrough}. The remaining claims follow, respectively, from Corollary \ref{corol:roughdistribmetric} and the definition of $d_F$.
\end{proof}

\begin{proof}[Proposition 2]
From the definition of $d_F$, it is easy to show that $E(d_F) = 1 - Rand_F$, where $Rand_F$ is the Rand index defined in \cite{frigui2007clustering}.
\end{proof}

\end{document}